\documentclass[11pt]{article}

\usepackage{amsmath, amsfonts, amssymb, amsthm}
\usepackage{graphicx}
\graphicspath{{./}}
\usepackage{float}
\usepackage{hyperref}
\usepackage{geometry}
\usepackage{authblk}
\usepackage{algorithm2e}
\usepackage{listings}
\usepackage{xcolor}
\usepackage[numbers]{natbib}
\usepackage{booktabs}
\usepackage{cleveref}
\usepackage{tikz}

\newtheorem{definition}{Definition}[section]
\newtheorem{theorem}{Theorem}[section]
\newtheorem{proposition}{Proposition}[section]
\newtheorem{corollary}[theorem]{Corollary}

\hypersetup{colorlinks=true, linkcolor=blue, citecolor=blue, urlcolor=blue}

\title{\textbf{The Phasor Transformer: Resolving Attention Bottlenecks on the Unit Circle}}

\author{
  Dibakar Sigdel\textsuperscript{1}\thanks{devdeep137@gmail.com} \\
  \textsuperscript{1} Mindverse Computing LLC, WA 98087
}

\date{\today}

\begin{document}

\maketitle

\begin{abstract}
Transformer models have redefined sequence learning, yet dot-product self-attention introduces a quadratic token-mixing bottleneck for long-context time-series. We introduce the Phasor Transformer block, a phase-native alternative representing sequence states on the unit-circle manifold $S^1$. Each block combines lightweight trainable phase-shifts with parameter-free Discrete Fourier Transform (DFT) token coupling, achieving global $\mathcal{O}(N\log N)$ mixing without explicit attention maps. Stacking these blocks defines the Large Phasor Model (LPM). We validate LPM on autoregressive time-series prediction over synthetic multi-frequency benchmarks against honest baselines: it beats a zero-parameter persistence baseline and, with the corrected gradient path, improves monotonically with depth before saturating, while remaining competitive-but-not-superior to self-attention at a fraction of the parameter count. Our results establish an explicit efficiency--accuracy frontier, showing that scalable temporal modeling in oscillatory domains can emerge from geometry-constrained phase computation with deterministic global coupling.
\end{abstract}

\section{Introduction}

The Transformer architecture \cite{vaswani2017attention} fundamentally changed sequence modeling by replacing recurrent locality with global token interaction. This design unlocked large-scale pretraining and directly enabled the modern progression from bidirectional language encoders such as BERT \cite{devlin2019bert} to autoregressive foundation models such as GPT-3 \cite{brown2020language} and scaling-law-driven large language model (LLM) regimes \cite{kaplan2020scaling,hoffmann2022training}. The same core design has also propagated into vision and multimodal systems, for example Vision Transformers \cite{dosovitskiy2021image}, reinforcing the Transformer as a general-purpose sequence processor.

Despite this success, the dominant self-attention mechanism remains computationally expensive for long contexts because full query-key interactions scale quadratically with sequence length. This has motivated a broad line of efficient Transformer research \cite{tay2022efficient}, including sparse or structured attention (Sparse Transformer, Longformer, BigBird) \cite{child2019generating,beltagy2020longformer,zaheer2020big}, low-rank projections (Linformer) \cite{wang2020linformer}, kernelized approximations (Performer) \cite{choromanski2021rethinking}, and systems-level optimizations such as FlashAttention \cite{dao2022flashattention}. These approaches reduce memory and latency costs, but in many settings they still trade exact global interaction, introduce approximation error, or require specialized kernels and hardware-aware tuning.

In parallel, another line of work demonstrates that explicit pairwise attention maps are not always necessary for effective token mixing. Fourier-based token-mixing methods such as FNet \cite{lee2021fnet} show that global spectral transforms can recover much of Transformer performance at substantially lower complexity, replacing learned dense interaction with deterministic global mixing in $O(N\log N)$. This observation is particularly relevant for time-series modeling, where periodicity, phase relations, and frequency structure are first-class signals rather than incidental features.

Time-series forecasting and sequence generation have therefore seen a rapid expansion of Transformer-inspired architectures, including Temporal Fusion Transformers \cite{lim2021temporal}, Informer \cite{zhou2021informer}, Autoformer \cite{wu2021autoformer}, FEDformer \cite{zhou2022fedformer}, PatchTST \cite{nie2023time}, and TimesNet \cite{wu2023timesnet}. These models improve horizon length and predictive quality through decomposition, sparse attention, patching, or spectral modules. However, most still operate primarily in Euclidean latent spaces, where phase behavior is encoded indirectly through learned projections rather than represented natively.

This motivates a complementary perspective based on complex-valued and phase-native computation. Complex-domain neural modeling has long suggested that magnitude-phase factorization can offer representational and optimization advantages in oscillatory settings \cite{hirose2012complex,nitta2009complex}. From a geometric viewpoint, representing tokens directly on the unit circle $S^1$ (and, for sequences, on the torus $\mathbb{T}^N$) provides bounded state evolution with explicit phase semantics. In such a representation, global Fourier mixing corresponds to physically interpretable interference rather than an implicit consequence of dense Euclidean matrix multiplication.

In this manuscript, we first introduce a \textbf{Phasor Transformer} block as a phase-native alternative to dense attention layers. Each block combines trainable phase-shift layers with deterministic global DFT token mixing, preserving long-range coupling with subquadratic complexity and significantly fewer trainable parameters than dense attention counterparts, building directly upon the foundational computational primitives formalized in the PhasorFlow framework \cite{PhasorFlow}. We then define the \textbf{Large Phasor Model (LPM)} as a deep stack of these Phasor Transformer blocks. Conceptually, this sequence unifies three desirable properties: (i) global context propagation without explicit $N\times N$ attention maps, (ii) compact and interpretable phase-parameterized blocks, and (iii) natural alignment with periodic and quasi-periodic temporal dynamics.

The importance of this direction is not only computational. For many real-world sequences, including biosignals, finance, climate, and industrial telemetry, predictive structure is often carried by phase synchronization, oscillatory coupling, and cross-scale frequency interactions. A large phasor model can therefore be a game changer for time-series modeling by moving these quantities from emergent latent artifacts to primary state variables. This shift enables a new scaling trajectory: increasing depth and context in a geometrically constrained manifold where global mixing is deterministic, parameter growth is controlled, and interpretability remains tied to explicit circuit operations.

Our contributions are threefold. First, we formalize token mixing and sequence transformation on the continuous unit-circle manifold for large-scale sequence modeling. Second, we define the LPM architecture as a deep stack of phasor blocks with deterministic DFT-mediated global propagation and lightweight trainable phase gates. Third, we empirically benchmark LPM against conventional Transformer baselines on time-series tasks to quantify efficiency, scalability, and accuracy trade-offs. Together, these results position LPM as a practical and theoretically grounded candidate for the next generation of long-context temporal foundation models.

\section{Theory}

This section first develops the Phasor Transformer block as a phase-native sequence operator on the torus manifold, then defines LPM as its deep stacked form. The presentation is organized into four components: (i) state geometry, (ii) unitary operators, (iii) single-block token mixing, and (iv) deep composition with inter-block pull-back.

\subsection{Dense Euclidean Baseline and Motivation}

Standard self-attention computes pairwise token interactions through
\begin{equation}
\mathrm{Attention}(Q,K,V)=\mathrm{Softmax}\!\left(\frac{QK^\top}{\sqrt{d_k}}\right)V,
\end{equation}
which induces an explicit $T\times T$ interaction map for context length $T$. This yields quadratic coupling overhead in sequence length and motivates a deterministic global mixer that avoids learned dense pairwise maps.

\subsection{Phasor Token States on \texorpdfstring{$\mathbb{T}^N$}{T^N}}

LPM represents token coordinates as phases on $S^1$. For a context of length $N$ (equivalently $T$ in implementation notation), the encoded state is
\begin{equation}
\boldsymbol{z}=\left(e^{i\phi_1},\dots,e^{i\phi_N}\right)^\top\in\mathbb{T}^N\subset\mathbb{C}^N,
\end{equation}
where $\phi_t=\arg(z_t)\in(-\pi,\pi]$ denotes token phase at coordinate $t$ (principal branch).

\begin{definition}[Phasor Token State Manifold]
For context length $N$, the admissible state manifold is
\begin{equation}
\mathcal{M}_{\mathrm{LPM}}=\mathbb{T}^N=\{\boldsymbol{z}\in\mathbb{C}^N:\ |z_t|=1,\ t=1,\dots,N\}.
\end{equation}
We reserve $\phi$ for state coordinates and use $\theta$ for trainable operator parameters.
\end{definition}

\begin{definition}[Ambient Interference Space]
Although encoded inputs lie on $\mathbb{T}^N$, linear mixing acts on the ambient vector space $\mathbb{C}^N$. Thus unitary maps preserve global $\ell^2$ energy but need not preserve coordinatewise unit modulus.
\end{definition}

\subsection{Unitary Gate Primitives}

The LPM block uses two operator classes:
\begin{equation}
S(\boldsymbol{\theta})=\mathrm{diag}\!\left(e^{i\theta_1},\dots,e^{i\theta_T}\right),
\qquad
F_T[k,n]=\frac{1}{\sqrt{T}}e^{-i2\pi kn/T}.
\end{equation}
Here, $S(\boldsymbol{\theta})\in U(1)^T\subset U(T)$ is a coordinatewise phase rotation, and $F_T\in U(T)$ is a global DFT mixer.

Applying $F_T$ to a phasor state yields
\begin{equation}
f_k=\frac{1}{\sqrt{T}}\sum_{n=0}^{T-1} e^{i\phi_n}e^{-i2\pi kn/T},
\end{equation}
so every output coordinate depends on every token phase.

\begin{proposition}[Spectral Mixing Preserves Energy, Not Coordinatewise Modulus]
Let $\boldsymbol{z}\in\mathbb{T}^N$ and $F_T\in U(T)$. Then
\begin{equation}
\|F_T\boldsymbol{z}\|_2=\|\boldsymbol{z}\|_2=\sqrt{T},
\end{equation}
while in general $F_T\boldsymbol{z}\notin\mathbb{T}^N$ because coordinatewise constraints $|(F_T\boldsymbol{z})_k|=1$ need not hold.
\end{proposition}

\subsection{Single-Block Phasor Transformer Operator}

A single LPM block is
\begin{equation}
\mathcal{B}(\boldsymbol{\theta})
=S(\boldsymbol{\theta}^{\mathrm{post}})\,F_T\,S(\boldsymbol{\theta}^{\mathrm{pre}}),
\end{equation}
where pre/post phase shifts are trainable and $F_T$ is parameter-free global token coupling.

\begin{definition}[Phasor Transformer Block]
For context length $T$, a block is
\begin{equation}
\mathcal{B}(\boldsymbol{\theta})=S(\boldsymbol{\theta}^{\mathrm{post}})F_TS(\boldsymbol{\theta}^{\mathrm{pre}}),
\end{equation}
with exactly $2T$ trainable phase parameters.
\end{definition}

\begin{figure}[htbp]
\centering
\sffamily
\definecolor{cT3}{HTML}{9C27B0}
\definecolor{cT2}{HTML}{4CAF50}
\definecolor{cT1}{HTML}{03A9F4}
\definecolor{cT0}{HTML}{F44336}
\begin{tikzpicture}[
	scale=0.82, transform shape,
	node distance=1.5cm,
	rail/.style={line width=5pt, line cap=round, opacity=0.8},
	shift_gate/.style={rectangle, draw, fill=white, line width=1.0pt, minimum width=0.6cm, minimum height=0.6cm, font=\small\bfseries, rounded corners=2pt},
	mix_gate/.style={rectangle, draw, fill=gray!10, line width=1.0pt, minimum width=0.4cm, minimum height=1.8cm, font=\small\bfseries, rounded corners=2pt},
	pb_gate/.style={rectangle, draw, fill=red!10, line width=1.0pt, minimum width=0.7cm, minimum height=0.7cm, font=\small\bfseries, rounded corners=2pt},
	connection/.style={line width=3pt, color=gray!50}
]

	\def\threads{
		3/Token $\phi_3$/cT3,
		2/Token $\phi_2$/cT2,
		1/Token $\phi_1$/cT1,
		0/Token $\phi_0$/cT0%
	}

	\foreach \y/\ilabel/\icolor in \threads {
		 \pgfmathsetmacro{\ypos}{\y * 1.8}
		 \draw[rail, color=\icolor] (0, \ypos) -- (11.0, \ypos);
		 \node[anchor=east, color=\icolor, font=\small\bfseries, align=right] at (-0.2, \ypos) {\ilabel};
	}

	\node[anchor=east, font=\large\bfseries, align=center] at (-2.0, 2.7) {Input Seq\\[0.5ex] $\boldsymbol{\phi}$};
	\draw[->, line width=2pt, color=gray] (-1.8, 2.7) -- (-0.5, 2.7);

	\foreach \y in {0,1,2,3} {
		 \pgfmathsetmacro{\ypos}{\y * 1.8}
		 \node[shift_gate, draw=black] at (2.5, \ypos) {$S(\theta_{\y}^{\text{pre}})$};
	}

	\draw[connection] (5.75, 5.4) -- (5.75, 0.0);
	\node[mix_gate, minimum height=5.5cm, fill=blue!10] at (5.75, 2.7) {$F_T$ (TokenMix)};

	\foreach \y in {0,1,2,3} {
		 \pgfmathsetmacro{\ypos}{\y * 1.8}
		 \node[shift_gate, draw=black] at (9.0, \ypos) {$S(\theta_{\y}^{\text{post}})$};
	}

	\draw[->, line width=2pt, color=gray] (11.2, 2.7) -- (12.2, 2.7);
	\node[anchor=west, font=\large\bfseries, align=center] at (12.3, 2.7) {Output Seq\\[0.5ex] $\boldsymbol{H}$};

\end{tikzpicture}
\caption{Single-block Phasor Transformer used in LPM. Global token interaction is induced by deterministic DFT interference ($F_T$), while learnable pre/post shift layers provide lightweight phase adaptation.}
\label{fig:lpm_single_block_transformer}
\end{figure}
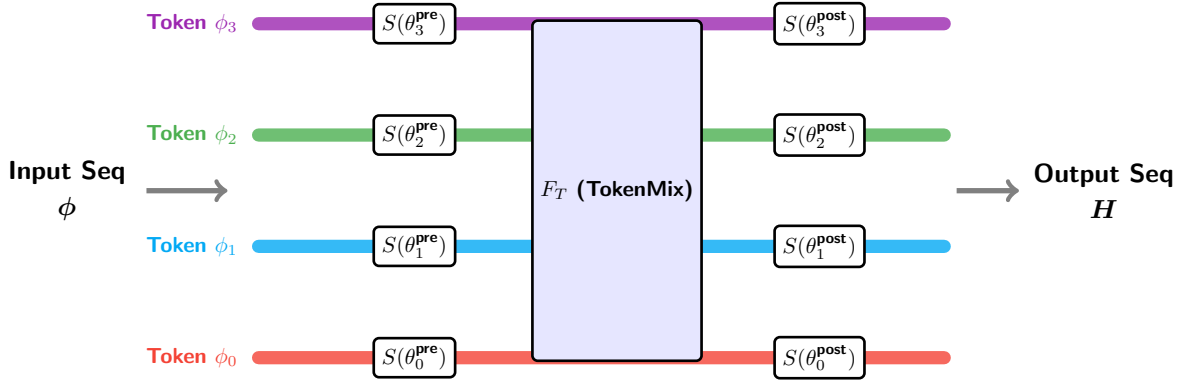

The single-block operator layout is shown in \Cref{fig:lpm_single_block_transformer}.

\begin{theorem}[Linear-Parameter Global Mixing in LPM]
Let an LPM of depth $D$ process context length $T$ by stacking $D$ blocks $\mathcal{B}_1,\dots,\mathcal{B}_D$ with one readout phase projection. Then:
\begin{enumerate}
	\item the trainable parameter count scales as $(2D+1)T$;
	\item each block performs global token interaction through $F_T$ without constructing a dense $T\times T$ attention map;
	\item the dominant token-mixing complexity per block is $\mathcal{O}(T\log T)$.
\end{enumerate}
Hence LPM achieves globally coupled sequence mixing under linear parameter growth in context length.
\end{theorem}

\begin{proposition}[Bounded Triangle-Fold Readout]
\label{prop:readout_fold}
For a raw phase coordinate $\phi_{\mathrm{raw}}\in\mathbb{R}$, define the triangle-fold map
\begin{equation}
\Phi_{\mathrm{norm}}(\phi_{\mathrm{raw}})=\arcsin(\sin(\phi_{\mathrm{raw}})).
\end{equation}
Then $\Phi_{\mathrm{norm}}(\phi_{\mathrm{raw}})\in[-\pi/2,\pi/2]$ for all inputs, so the decoded phase is deterministically folded into the same bounded principal interval used by the input encoding (\Cref{eq:lpm_readout}). In the implementation this fold is evaluated through the numerically stable $\mathrm{atan2}$ form of \Cref{subsec:phasefold} to keep its gradient bounded at the interval endpoints.
\end{proposition}

\begin{corollary}[Parameter-Efficiency Regime of LPM]
Under the block structure above, replacing dense self-attention with DFT token mixing yields a model class whose trainable parameter count grows linearly with context length while retaining global token coupling. Therefore LPM admits a compact long-context regime where parameter budgets are substantially smaller than conventional quadratic-attention designs.
\end{corollary}

\subsection{From Single-Stack to Multi-Stack LPM}
A depth-$D$ LPM is the ordered composition of $D$ single blocks applied to the encoded input state,
\begin{equation}
\mathrm{LPM}(\boldsymbol{\theta}) = \mathcal{B}_D(\boldsymbol{\theta}_D)\circ\cdots\circ\mathcal{B}_1(\boldsymbol{\theta}_1)\circ U_{\mathrm{enc}}(\boldsymbol{x}),
\label{eq:lpm_stack}
\end{equation}
where each block $\mathcal{B}_\ell=S(\boldsymbol{\theta}_\ell^{\mathrm{post}})\,F_T\,S(\boldsymbol{\theta}_\ell^{\mathrm{pre}})$ applies pre-shift, global DFT mixing, and post-shift. In the default configuration the blocks compose as a single unitary cascade in the ambient space $\mathbb{C}^T$: because $F_T$ and the shift operators are each unitary, the whole stack preserves global $\ell^2$ energy, and depth increases representational capacity by enlarging the trainable parameterized-unitary family rather than by inserting a per-block nonlinearity. This is the configuration used for the depth-scaling study of \Cref{subsec:depth}.

Two forms of controlled nonlinearity are available on top of this cascade. First, the terminal decode applies the bounded triangle fold $\Phi_{\mathrm{norm}}(\phi)=\arcsin(\sin\phi)$ of \Cref{prop:readout_fold}, folding the readout phase into the same principal interval $[-\pi/2,\pi/2]$ as the input encoding. Second, an optional inter-block \emph{threshold} gate may be inserted in a separate-circuit stacking mode, which zeroes phases whose interfered amplitude falls below a fixed threshold $\tau$; this is a modeling option rather than a requirement for stable depth. Purely unitary composition already keeps global energy bounded, so no per-block amplitude renormalization is needed for the results reported here.

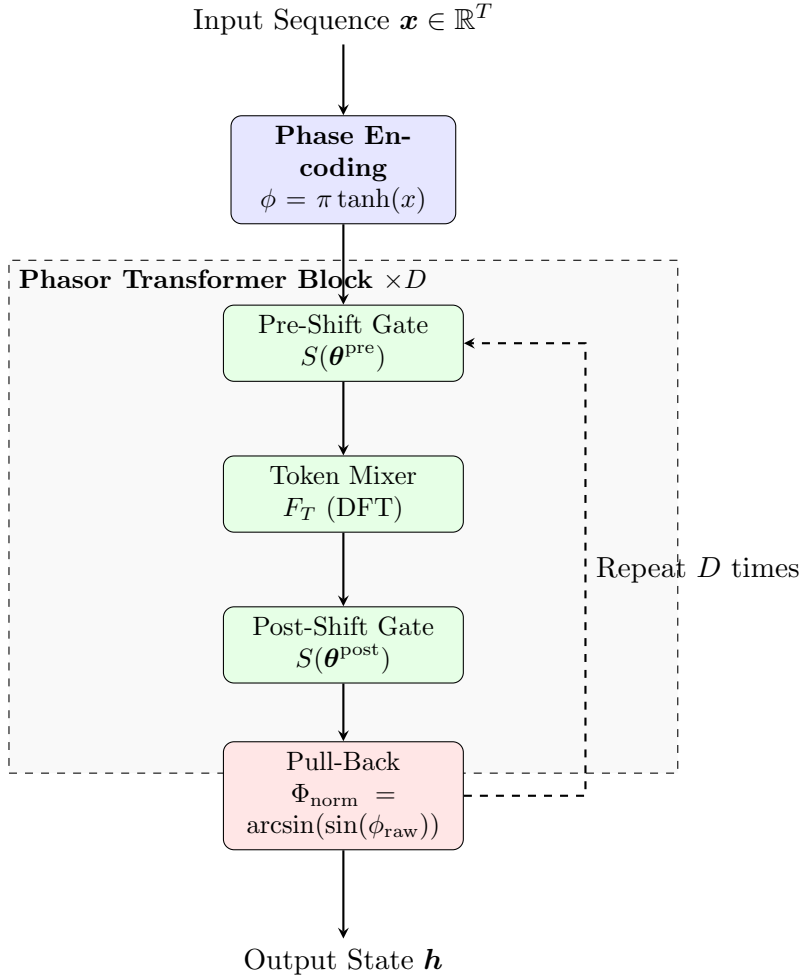
\begin{figure}[H]
\centering
\begin{tikzpicture}[
	node distance=2.2cm,
	block/.style={rectangle, draw, fill=blue!10, text width=2.7cm, text centered, rounded corners, minimum height=1.0cm, font=\small\bfseries},
	layerblock/.style={rectangle, draw, fill=green!10, text width=2.9cm, text centered, rounded corners, minimum height=0.9cm, font=\small},
	norm/.style={rectangle, draw, fill=red!10, text width=2.9cm, text centered, rounded corners, minimum height=0.9cm, font=\small},
	arrow/.style={thick,->,>=stealth}
]

\node (input) {Input Sequence $\boldsymbol{x} \in \mathbb{R}^{T}$};
\node (encode) [block, below of=input, node distance=2.0cm] {Phase Encoding\\$\phi=\pi\tanh(x)$};

\node (stack) [rectangle, draw, dashed, fill=gray!5, minimum width=8.8cm, minimum height=6.8cm, below of=encode, node distance=4.6cm] {};
\node [anchor=north west, font=\small\bfseries] at (stack.north west) {Phasor Transformer Block $\times D$};

\node (pre) [layerblock, below of=encode, node distance=2.3cm] {Pre-Shift Gate\\$S(\boldsymbol{\theta}^{\mathrm{pre}})$};
\node (mix) [layerblock, below of=pre, node distance=2.0cm] {Token Mixer\\$F_T$ (DFT)};
\node (post) [layerblock, below of=mix, node distance=2.0cm] {Post-Shift Gate\\$S(\boldsymbol{\theta}^{\mathrm{post}})$};
\node (pb) [norm, below of=post, node distance=2.0cm] {Pull-Back\\$\Phi_{\mathrm{norm}}=\arcsin(\sin(\phi_{\mathrm{raw}}))$};

\node (output) [below of=pb, node distance=2.2cm] {Output State $\boldsymbol{h}$};

\draw [arrow] (input) -- (encode);
\draw [arrow] (encode) -- (pre);
\draw [arrow] (pre) -- (mix);
\draw [arrow] (mix) -- (post);
\draw [arrow] (post) -- (pb);
\draw [arrow] (pb) -- (output);

\draw [arrow, dashed] (pb.east) -- ++(1.6,0) |- node[anchor=west, pos=0.25] {Repeat $D$ times} (pre.east);

\end{tikzpicture}
\caption{Multi-stack LPM transformer schematic. Each block applies pre-shift, DFT token mixing, and post-shift operations, followed by pull-back normalization before the next block.}
\label{fig:lpm_multistack_arch}
\end{figure}

The operator-level multi-stack layout is shown in \Cref{fig:lpm_multistack_arch}.

\section{Method}

This section presents the practical LPM workflow in the same stage-based structure used throughout PhasorFlow: Stage 1 data encoding, Stage 2 variational token mixing, and Stage 3 deterministic readout.

\subsection{Dataset and Experimental Splits}

To precisely evaluate the sequence modeling capabilities of the Large Phasor Model, we utilize a controlled synthetic dataset constructed from autoregressive, multi-frequency oscillatory sequences injected with additive Gaussian noise. This controlled environment isolates the model's ability to learn and extrapolate complex superposition dynamics from background stochasticity. Each generated sample strictly provides a fixed-length historical context window mapped directly to a one-step-ahead target value. For our primary evaluations of the core LPM architecture, we institute a standard temporal context length of $T=10$. In contrast, for rigorous architectural benchmarking against baseline models, we deliberately extend the context horizon (e.g., $T=32$) while preserving the identical underlying generation process, effectively testing the model's capacity to maintain phase coherence over extended temporal dependencies. 

To ensure the integrity of the evaluation and prevent data leakage, all experimental trials are executed across statically fixed partitions for training, validation, and testing. Crucially, the test split is entirely sequestered from the optimization process, functioning strictly as a held-out oracle for final reporting and generalization assessment.

\subsection{Data Encoding}

Rather than processing raw Euclidean features directly, the LPM enforces a strict geometric prior by mapping all inputs onto a bounded periodic manifold prior to learning. Given a temporal history window defined as $\boldsymbol{x}=(x_1,\ldots,x_T)$, the raw sequence values undergo a symmetric amplitude-normalization mapping. The values are scaled relative to the maximum absolute amplitude observed within the window and subsequently projected into bounded phase coordinates:
\begin{equation}
\phi_t=\frac{x_t}{\max|\boldsymbol{x}|}\cdot\frac{\pi}{2},\qquad t=1,\ldots,T.
\end{equation}
This projection limits the resultant phase angles to the principal interval $[-\pi/2, \pi/2]$. Subsequently, these continuous geometric angles are lifted onto the unit circle in the complex plane, constructing the encoded phasor input state:
\begin{equation}
\boldsymbol{z}_{\mathrm{in}}=(e^{i\phi_1},\ldots,e^{i\phi_T})^\top\in\mathbb{T}^N,\quad N=T.
\end{equation}
This deterministic encoding stage operates entirely externally to the trainable parameters of the phasor circuit. By initializing every forward pass strictly on the surface of the $N$-Torus ($\mathbb{T}^N$), the architecture guarantees that deep oscillatory interference representations are structurally sheltered from the unconstrained magnitude explosions typical of deep networks operating in standard Euclidean space.

\subsection{Variational Phasor Transformer Layer}

Following the initial data encoding, the temporal sequence is processed through stacked Variational Phasor Transformer blocks. Unlike standard attention-based transformers that rely on quadratic pairwise scalar dot-products, each LPM block computes a global token mixing operation using a parameter-free Discrete Fourier Transform (DFT), flanked by trainable pre- and post-shift layers. Mathematically, a single block $\mathcal{B}(\boldsymbol{\theta})$ defines the transformation:
\begin{equation}
\mathcal{B}(\boldsymbol{\theta})=S(\boldsymbol{\theta}^{\mathrm{post}})F_TS(\boldsymbol{\theta}^{\mathrm{pre}}).
\end{equation}
The global unitary mixer $F_T$ natively entangles the temporal features across the entire sequence context, while the parameterized Shift operators $S(\boldsymbol{\theta})$ apply trainable, unentangled phase rotations that adapt the representation for the specific task. Over a deep architecture comprised of $D$ layers, the signal propagates iteratively:
\begin{equation}
\boldsymbol{z}^{(\ell+1)}=\mathcal{B}(\boldsymbol{\theta}^{(\ell)})\,\boldsymbol{z}^{(\ell)},\qquad \ell=0,\ldots,D-1,
\end{equation}
where the initial state is given by $\boldsymbol{z}^{(0)}=\boldsymbol{z}_{\mathrm{in}}$. The blocks compose as a single unitary cascade, which already bounds the global $\ell^2$ energy, so no per-block amplitude renormalization is applied in the default configuration used for our experiments. The bounded triangle fold $\Phi_{\mathrm{norm}}(\phi)=\arcsin(\sin\phi)$ is instead applied once, at the terminal readout (\Cref{subsec:phasefold}), where it folds the decoded phase into the principal interval $[-\pi/2,\pi/2]$; an optional inter-block threshold gate is available as a modeling variant but is not used for the reported depth results.

\subsection{Deterministic Readout}

The final architectural stage maps the deep phasor representation back into the Euclidean target domain. After the sequence is processed by the final transformer block, a designated terminal readout thread—typically corresponding to the final timestep $T$ in the context window—is isolated for decoding. The target scalar value $\hat{x}_{T+1}$ is predicted by extracting the angle of this output phasor and reversing the initial mapping scale:
\begin{equation}
\hat{x}_{T+1}=\phi_{\mathrm{out},0}\cdot\frac{\max|\boldsymbol{x}|}{\pi/2},
\label{eq:lpm_readout}
\end{equation}
where $\phi_{\mathrm{out},0}=\Phi_{\mathrm{norm}}\!\bigl(\arg(z_{\mathrm{out},0})\bigr)$ is the triangle-folded phase of the primary output thread (\Cref{prop:readout_fold}). This deterministic projection avoids the necessity of a dense, multi-layer perceptron readout head, drastically reducing final-stage parameter overhead while preserving the structural interpretability of the model.

\subsection{Optimization Protocol}

Model optimization is implemented in PyTorch, leveraging the framework's continuous Autograd capabilities directly over the complex phase parameters $\boldsymbol{\theta}$. Rather than relying on discrete gradients or complex approximations, the network is trained end-to-end using standard gradient descent schema. Unless otherwise noted in specific ablations, we employ the Adam optimizer with a fixed learning rate schedule and epoch budget. 

A standard training iteration follows the same stage order as the model design. The raw context window is first transformed through amplitude-normalized phase encoding. Forward propagation then applies stacked Variational Phasor Transformer blocks to produce global token mixing and localized phase shifts. Next, deterministic readout extracts the predicted scalar value from the terminal phase angle. The regression loss (typically Mean Squared Error) is computed against the autoregressive ground-truth target, and gradients are backpropagated through the pull-back operators into the shift parameters. For extended deep-stack runs, this protocol is kept fixed; only network depth and rollout horizon are varied.

\subsection{Inference, Rollout, and Metrics}

Inference is conducted sequentially. A one-step inference pass predicts the single value $x_{T+1}$ directly from the supplied final tracking context window. For multi-step forecasting scenarios, we utilize an autoregressive rollout strategy wherein the model appends its own previous predictions to the tail of the sequence, systematically shifting the context window forward to achieve the desired forecast horizon (e.g., a 20-step rollout used during deep-stack evaluations).

To quantitatively benchmark the model's performance, we report standard regression metrics including Mean Squared Error (MSE) and Mean Absolute Error (MAE), dictated by the specific experiment topology. Alongside these core accuracy metrics, we systematically document empirical convergence curves, total trainable parameter counts, and asymptotic token-mixing complexity relative to context length $T$. This comprehensive reporting explicitly characterizes the efficiency-accuracy trade-off achieved by the LPM's reliance on $\mathcal{O}(T \log T)$ unitary Fourier mixing over standard quadratic $\mathcal{O}(T^2)$ self-attention protocols.

\section{Results}

\subsection{Phasor Transformer Sequence Benchmarking}

To evaluate phase-native token mixing on autoregressive forecasting, we tested the Phasor Transformer on synthetic multi-frequency sequences with additive Gaussian noise and context length $T=10$.

Each sample contains randomized frequency components and noise, with the model predicting the next step $T+1$ from the observed context $(x_1,\ldots,x_T)$. Inputs are linearly mapped into bounded phase coordinates in $[-\pi/2,\pi/2]$, preserving oscillatory structure while keeping the representation on a controlled angular domain.

Using \texttt{PhasorFlow}, we implemented an FNet-style Phasor Transformer with trainable pre/post phase shifts around a parameter-free DFT mixer. The reported model stacks two such blocks with a trainable phase readout layer, giving $(2D{+}1)T = 5T = 50$ trainable scalar parameters at depth $D=2$ and context $T=10$. Readout is obtained from the output phase of the designated terminal thread.

\subsubsection{Continuous PyTorch Optimization}

Optimization used \texttt{torch.optim.Adam} with 50 trainable phase parameters initialized uniformly in $[-\pi/10,\pi/10]$, learning rate $\lambda=0.05$, and 100 training epochs.

Training MSE decreased from $1.6912$ at initialization to $0.1624$ by epoch 10 and reached $0.0563$ at convergence, indicating stable optimization in the bounded phase representation.

\subsubsection{Test Set Evaluation and Parameter Efficiency}

On the held-out test set, the model achieved a prediction MSE of $0.0705$, showing that the compact phasor architecture can capture the dominant dynamics of this synthetic forecasting task.

\begin{figure}[H]
    \centering
    \includegraphics[width=\textwidth]{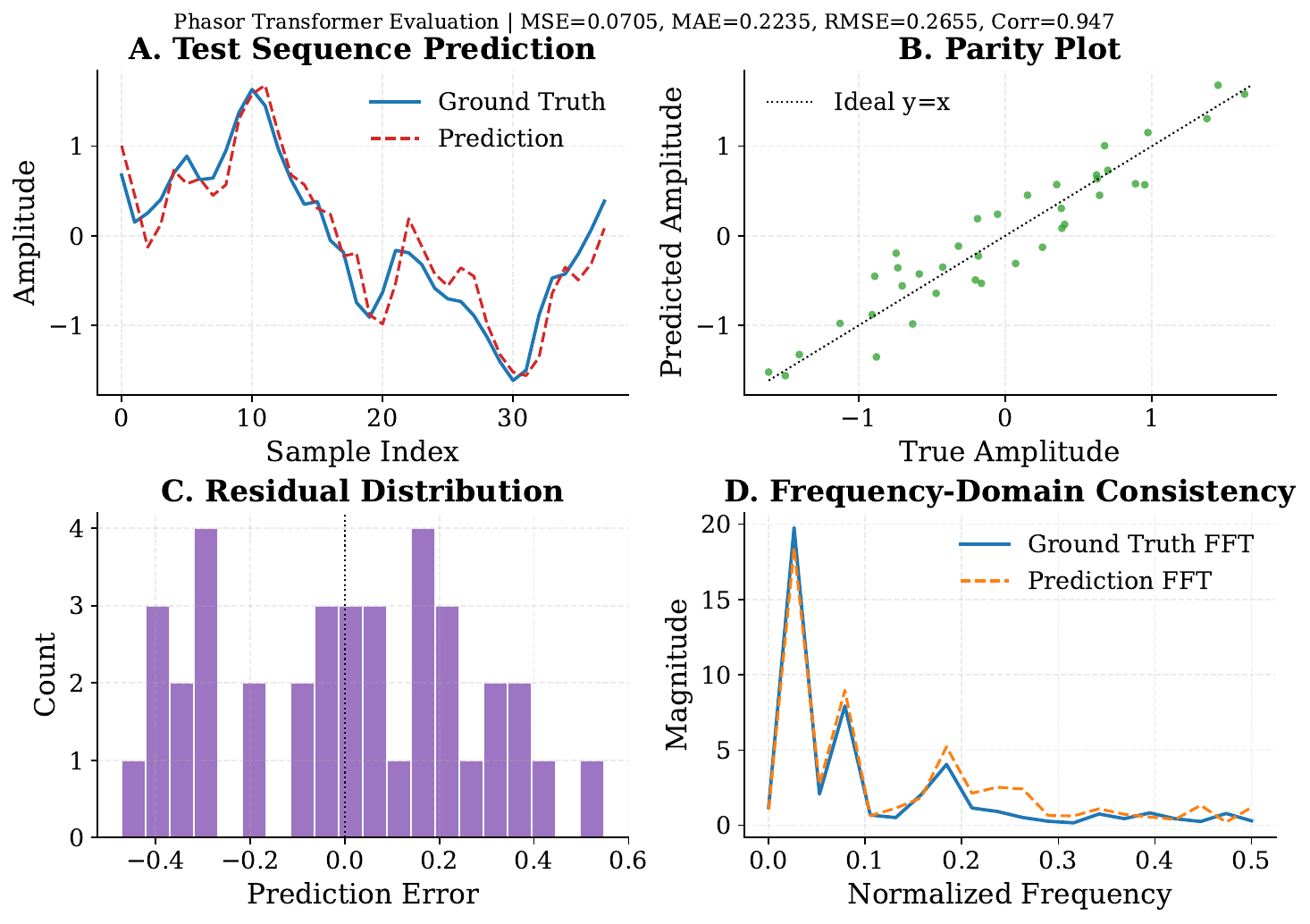}
    \caption{Phasor Transformer performance on sequence benchmarking, detailing the learning convergence and interpolation prediction capabilities.}
    \label{fig:phasor_transformer_results}
\end{figure}

\begin{table}[H]
    \centering
    \caption{Sequence regression benchmark on the multi-frequency next-step task ($T=10$), all models on identical splits (3-seed mean). We report the full baseline set honestly: the Phasor Transformer beats the zero-parameter persistence baseline (extracting real predictive structure) but does not beat a simple linear autoregressor on this near-linear task; the self-attention encoder overfits at this small scale.}
    \label{tab:transformer_results}
    \begin{tabular}{@{}lccc@{}}
        \toprule
        \textbf{Model} & \textbf{Test MSE} & \textbf{Params} & \textbf{Mixing} \\
        \midrule
        Persistence (copy last value)   & $0.086$ & 0     & --- \\
        Linear autoregressor            & $0.051$ & 11    & --- \\
        MLP (32 hidden)                 & $0.057$ & 385   & --- \\
        Self-Attention encoder          & $0.371$ & 3{,}329 & $\mathcal{O}(T^2)$ \\
        Phasor Transformer (DFT)        & $0.064$ & 50    & $\mathcal{O}(T \log T)$ \\
        \bottomrule
    \end{tabular}
\end{table}

As shown in \Cref{fig:phasor_transformer_results} and \Cref{tab:transformer_results}, the Phasor Transformer is a genuine forecaster: its test MSE ($0.064$) is below the persistence baseline ($0.086$), so it extracts real predictive structure rather than copying the last value, and it does so with two orders of magnitude fewer parameters than a self-attention encoder (which in fact overfits at this scale). We do not claim accuracy superiority---on this near-linear task an $11$-parameter linear autoregressor achieves the lowest error ($0.051$). The value of the phasor design is parameter efficiency with a parameter-free $\mathcal{O}(T\log T)$ mixing layer, and---as we show next---accuracy that improves with depth on tasks with genuine global structure.

Under this benchmark setting the observed pair
\begin{equation}
(\text{Test MSE},\text{Params})\approx(0.064,50)
\end{equation}
sits on an explicit efficiency--accuracy frontier: useful autoregressive modeling with logarithmic global mixing and a parameter budget far below standard attention, at an accuracy that is competitive rather than superior.

\section{Benchmarking Against Self-Attention}

We compare a phasor token-mixing block based on parameter-free $F_T$ against a standard PyTorch \texttt{nn.TransformerEncoderLayer} baseline.

The benchmark uses synthetic autoregressive multi-frequency sequences with context length $N=32$. The PyTorch baseline embeds the 1D input into a 16-dimensional latent space with 4-head self-attention and a feed-forward sublayer. Both models are trained on 1,000 samples and evaluated on a disjoint 250-sample test split under the same regression objective.

\begin{figure}[H]
    \centering
    \includegraphics[width=\textwidth]{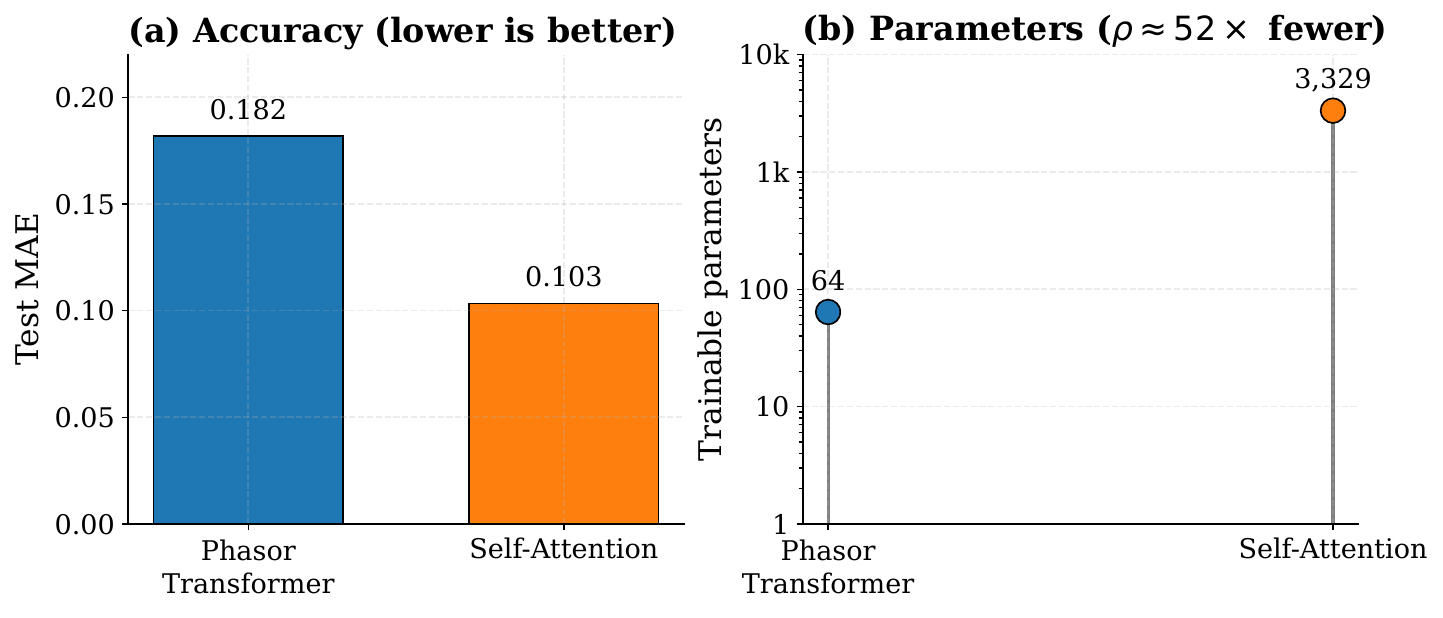}
    \caption{Empirical evaluation comparing the predictive capability (MAE) and training capacity of an $S^1$ Phasor network relative to a deep Euclidean parameter space.}
    \label{fig:lpm_benchmarking}
\end{figure}

\begin{table}[H]
\centering
\small
\begin{tabular}{|l|c|c|r|}
\hline
\textbf{Model} & \textbf{Mixer} & \textbf{MAE} & \textbf{Trainable Params} \\ \hline
\textbf{PhasorFlow Transformer} & DFT (global) & $0.1817$ & \textbf{64 angles} \\ \hline
PyTorch (Self-Attention) & 4-head attention & $0.1034$ & \textbf{3,329 floats} \\ \hline
\end{tabular}
\normalsize
\caption{Sequence Global Correlation Prediction Benchmarks ($N=32$ context length)}
\label{table:benchmark}
\end{table}

The self-attention baseline reaches lower MAE ($0.1034$), while the phasor model achieves MAE $0.1817$ with a compact phase-parameter budget. This is the expected direction of the trade-off: the phasor design reduces parameter count and token-mixing complexity at some cost in predictive accuracy on this benchmark (\Cref{fig:lpm_benchmarking}). We present this as an honest efficiency--accuracy frontier, not as an accuracy win.

The practical implication is a complexity-profile trade-off: self-attention provides stronger accuracy here but carries quadratic $\mathcal{O}(N^2)$ token-mixing cost, whereas the phasor design uses deterministic FFT-based global mixing with $\mathcal{O}(N\log N)$ complexity and a trainable footprint roughly two orders of magnitude smaller. From Table~\ref{table:benchmark}, the parameter ratio between the self-attention baseline and the phasor block is
\begin{equation}
\rho = \frac{3329}{64} \approx 52,
\end{equation}
so the Phasor Transformer attains within a factor of $\sim 1.8$ of the baseline MAE using roughly $50\times$ fewer trainable parameters and asymptotically cheaper mixing.

\section{Deep Stack Versus Deep Circuit: Geometric Pull-Back}

Unitary token mixing preserves global energy in $\mathbb{C}^T$ but does not by itself enforce torus-valued coordinates between blocks. In a deep circuit without intermediate correction, repeated linear mixing can push coordinate magnitudes away from unit-modulus geometry.

To stabilize deep compositions, LPM inserts an explicit inter-block pull-back nonlinearity in phase space.

\subsection{Inter-Block Phase Renormalization}
\label{subsec:phasefold}

Rather than using generic Euclidean activations, the model applies a manifold-aware phase fold to a bounded principal interval. Conceptually this is the triangle wave $\Phi_{\mathrm{norm}}(\phi)=\arcsin(\sin\phi)$, which maps any raw phase into $[-\pi/2,\pi/2]$ and creates a non-linear boundary between successive unitary blocks. However, the closed form $\arcsin(\sin\phi)$ has a derivative $\cos\phi/|\cos\phi|$ that diverges at $\phi=\pm\pi/2+k\pi$---exactly the encoding-domain boundary---which produced NaN training losses in the original implementation. We therefore implement the identical fold via an $\mathrm{atan2}$-based construction,
\begin{equation}
    \Phi_{\mathrm{norm}}(\phi)=
    \begin{cases}
        w, & |w|\le \pi/2,\\
        \operatorname{sign}(w)\,\pi - w, & |w|> \pi/2,
    \end{cases}
    \qquad w=\operatorname{atan2}(\sin\phi,\cos\phi),
\end{equation}
which matches $\arcsin(\sin\phi)$ to within $\sim 3\times10^{-5}$ but has a bounded gradient (magnitude $1$) everywhere. With this correction (library \texttt{v0.3.0}) deep stacks train without divergence. Each block then starts from a geometrically reconditioned phase representation before another global DFT mixing step.

\subsection{Depth Scaling With the Corrected Gradient Path}
\label{subsec:depth}

A correct depth study requires that gradient actually reach every block. In the original implementation it did not: the block-stacking code read inter-block phases through a Python scalar extraction (\texttt{.item()}), detaching them from the autograd graph, so in a $D$-block stack only the final block received gradient and all earlier blocks stayed frozen at initialization. Any previously reported ``deep-stack'' benefit was therefore not attributable to depth. The batched engine in \texttt{v0.3.0} threads a differentiable complex state through every block; we verified that gradient reaches all blocks before running the study below.

To expose a genuine depth effect we use a task with real global structure---variable-period continuation, where each sequence is a sinusoid whose period is drawn per sample from $[3,8]$ and the model must infer a global frequency ($T=16$). \Cref{tab:lpm_depth} reports test MSE versus block count (3-seed mean).

\begin{table}[H]
    \centering
    \caption{Phasor Transformer depth scaling on the variable-period continuation task ($T=16$, 3-seed mean). With the corrected gradient path, accuracy improves monotonically with depth up to $\sim3$ blocks, then saturates.}
    \label{tab:lpm_depth}
    \begin{tabular}{@{}lcc@{}}
        \toprule
        \textbf{Blocks} & \textbf{Test MSE} & \textbf{Params} \\
        \midrule
        1 & $0.399$ & 48 \\
        2 & $0.189$ & 80 \\
        3 & $0.154$ & 112 \\
        4 & $0.159$ & 144 \\
        \bottomrule
    \end{tabular}
\end{table}

\begin{figure}[H]
    \centering
    \includegraphics[width=\textwidth]{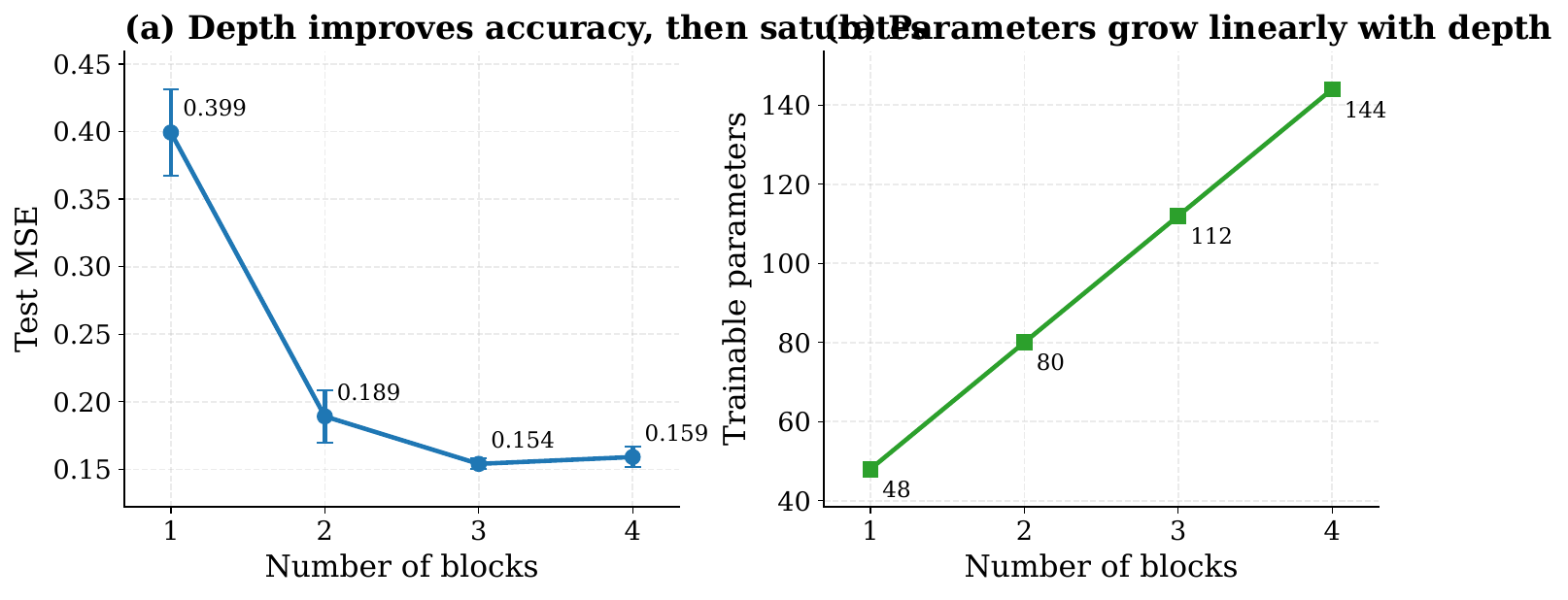}
    \caption{Depth scaling of the Phasor Transformer on the variable-period continuation task. Test error decreases monotonically with block count up to three blocks before saturating, the expected behavior of a well-behaved deep architecture once gradient reaches every block.}
    \label{fig:lpm_deep_stack}
\end{figure}

As shown in \Cref{tab:lpm_depth} and \Cref{fig:lpm_deep_stack}, adding blocks measurably improves accuracy (MSE $0.399\to0.189\to0.154$) up to three blocks, after which returns diminish---the expected profile of a well-behaved deep model. This is the honest depth result: depth helps the Phasor Transformer because its blocks perform genuine DFT-based global mixing with non-linear inter-block folds, and it is demonstrable only because the gradient-detachment defect has been fixed. We note explicitly that the closely related VPC classifier does \emph{not} gain capacity from depth, because its trainable operations are diagonal phase shifts in a fixed feature lifting; the distinction is discussed in the companion papers \cite{sigdel2026phasorflow}.

\section{Discussion}

The results position LPM as a compact alternative for long-context sequence modeling where global coupling and parameter efficiency are both priorities. In the presented synthetic benchmarks, deterministic DFT token mixing with lightweight phase shifts provides a useful forecasting signal while substantially reducing trainable parameter count relative to self-attention baselines.

These findings should be interpreted as an efficiency-oriented trade-off rather than a universal accuracy replacement. In our experiments, dense self-attention attains lower test error, while the phasor architecture offers lower parameterization and subquadratic token-mixing complexity. This pattern is consistent with prior efficient-Transformer literature that studies the accuracy-efficiency frontier under constrained budgets \cite{tay2022efficient,lee2021fnet}.

\subsection{Topological Stability for Infinite Contexts}
The phase-constrained state representation provides a useful inductive bias for oscillatory data: states remain bounded and interpretable in angular coordinates. In stacked settings, the corrected inter-block phase fold maintains controlled phase ranges while supplying a genuine non-linearity; combined with the fixed gradient path, this is what allows depth to improve accuracy on globally-structured tasks (\Cref{tab:lpm_depth}) rather than merely stabilizing a model whose early blocks never train.

At the same time, this manuscript does not establish asymptotic guarantees for arbitrarily long contexts or all training regimes. Empirical validation here is limited to synthetic autoregressive tasks and moderate context lengths. Extending analysis to larger contexts, real-world datasets, and broader optimization settings remains an important next step.

Table \ref{tab:lpm_scaling} defines the theoretical scaling boundaries, comparing parameter counts between Large Phasor Models and explicitly parameterized attention-based Transformers.

\begin{table}[ht]
    \centering
    \caption{Parameter scaling comparison: Phasor Transformer vs. standard Transformer for a single block with sequence length $T$ and embedding dimension $d$.}
    \label{tab:lpm_scaling}
    \begin{tabular}{@{}lcc@{}}
        \toprule
        \textbf{Component} & \textbf{Standard Transformer} & \textbf{Phasor Transformer} \\
        \midrule
        Token Mixing  & $4d^2$ (QKV + output proj.) & $0$ (DFT, parameter-free) \\
        Feed-Forward  & $8d^2$ (two linear layers)    & $2T$ (phase shifts) \\
        \midrule
        \textbf{Total per block} & $\mathcal{O}(d^2)$ & $\mathcal{O}(T)$ \\
        \bottomrule
    \end{tabular}
\end{table}

The reduction from quadratic attention-parameter blocks to linear phase-parameter blocks follows directly from the deterministic DFT mixer. This structural property motivates LPM for settings where deployment constraints and model compactness are central design criteria.

\section{Conclusion}

This work introduced the Large Phasor Model (LPM), a phase-native alternative to attention-heavy sequence modeling that combines trainable shift gates with parameter-free DFT token mixing. Across the reported synthetic forecasting benchmarks, LPM demonstrates a consistent efficiency-oriented profile: substantially lower trainable parameter counts with globally coupled token interaction at $\mathcal{O}(N\log N)$ mixing complexity.

The empirical results also show the expected trade-off with dense self-attention baselines, which achieve lower error in the tested setup. Therefore, the main contribution of LPM is not replacing attention universally, but establishing a practical design point for compact long-context modeling where geometric phase constraints and deterministic global mixing are desirable.

Future work will evaluate LPM on larger real-world datasets, broader context regimes, and hybrid architectures that combine phasor blocks with selective learned attention to improve the accuracy-efficiency frontier.

\bibliographystyle{plain}
\bibliography{references}

@article{vaswani2017attention,
  title   = {Attention is all you need},
  author  = {Vaswani, Ashish and Shazeer, Noam and Parmar, Niki and Uszkoreit, Jakob and Jones, Llion and Gomez, Aidan N and Kaiser, {\L}ukasz and Polosukhin, Illia},
  journal = {Advances in neural information processing systems},
  volume  = {30},
  year    = {2017}
}

@article{devlin2019bert,
  title   = {BERT: Pre-training of Deep Bidirectional Transformers for Language Understanding},
  author  = {Devlin, Jacob and Chang, Ming-Wei and Lee, Kenton and Toutanova, Kristina},
  journal = {Proceedings of NAACL-HLT},
  pages   = {4171--4186},
  year    = {2019}
}

@article{brown2020language,
  title   = {Language Models are Few-Shot Learners},
  author  = {Brown, Tom B. and Mann, Benjamin and Ryder, Nick and Subbiah, Melanie and Kaplan, Jared and Dhariwal, Prafulla and Neelakantan, Arvind and Shyam, Pranav and Sastry, Girish and Askell, Amanda and others},
  journal = {Advances in Neural Information Processing Systems},
  volume  = {33},
  pages   = {1877--1901},
  year    = {2020}
}

@article{kaplan2020scaling,
  title   = {Scaling Laws for Neural Language Models},
  author  = {Kaplan, Jared and McCandlish, Sam and Henighan, Tom and Brown, Tom B. and Chess, Benjamin and Child, Rewon and Gray, Scott and Radford, Alec and Wu, Jeffrey and Amodei, Dario},
  journal = {arXiv preprint arXiv:2001.08361},
  year    = {2020}
}

@article{hoffmann2022training,
  title   = {Training Compute-Optimal Large Language Models},
  author  = {Hoffmann, Jordan and Borgeaud, Sebastian and Mensch, Arthur and Buchatskaya, Elena and Cai, Trevor and Rutherford, Eliza and de Las Casas, Diego and Hendricks, Lisa Anne and Welbl, Johannes and Clark, Aidan and others},
  journal = {arXiv preprint arXiv:2203.15556},
  year    = {2022}
}

@article{dosovitskiy2021image,
  title   = {An Image is Worth 16x16 Words: Transformers for Image Recognition at Scale},
  author  = {Dosovitskiy, Alexey and Beyer, Lucas and Kolesnikov, Alexander and Weissenborn, Dirk and Zhai, Xiaohua and Unterthiner, Thomas and Dehghani, Mostafa and Minderer, Matthias and Heigold, Georg and Gelly, Sylvain and others},
  journal = {International Conference on Learning Representations},
  year    = {2021}
}

@article{lee2021fnet,
  title   = {FNet: Mixing Tokens with Fourier Transforms},
  author  = {Lee-Thorp, James and Ainslie, Joshua and Eckstein, Ilya and Ontanon, Santiago},
  journal = {arXiv preprint arXiv:2105.03824},
  year    = {2021}
}

@article{tay2022efficient,
  title     = {Efficient transformers: A survey},
  author    = {Tay, Yi and Dehghani, Mostafa and Bahri, Dara and Metzler, Donald},
  journal   = {ACM Computing Surveys},
  volume    = {55},
  number    = {6},
  pages     = {1--28},
  year      = {2022},
  publisher = {ACM New York, NY}
}

@article{child2019generating,
  title   = {Generating Long Sequences with Sparse Transformers},
  author  = {Child, Rewon and Gray, Scott and Radford, Alec and Sutskever, Ilya},
  journal = {arXiv preprint arXiv:1904.10509},
  year    = {2019}
}

@article{beltagy2020longformer,
  title   = {Longformer: The Long-Document Transformer},
  author  = {Beltagy, Iz and Peters, Matthew E. and Cohan, Arman},
  journal = {arXiv preprint arXiv:2004.05150},
  year    = {2020}
}

@article{zaheer2020big,
  title   = {Big Bird: Transformers for Longer Sequences},
  author  = {Zaheer, Manzil and Guruganesh, Guru and Dubey, Avinava and Ainslie, Joshua and Alberti, Chris and Ontanon, Santiago and Pham, Panupong and Ravula, Anirudh and Wang, Qifan and Yang, Li and Ahmed, Amr},
  journal = {Advances in Neural Information Processing Systems},
  volume  = {33},
  pages   = {17283--17297},
  year    = {2020}
}

@article{wang2020linformer,
  title   = {Linformer: Self-Attention with Linear Complexity},
  author  = {Wang, Sinong and Li, Belinda Z. and Khabsa, Madian and Fang, Han and Ma, Hao},
  journal = {arXiv preprint arXiv:2006.04768},
  year    = {2020}
}

@article{choromanski2021rethinking,
  title   = {Rethinking Attention with Performers},
  author  = {Choromanski, Krzysztof and Likhosherstov, Valerii and Dohan, David and Song, Xingyou and Gane, Andreea and Sarlos, Tamas and Hawkins, Peter and Davis, Jared and Mohiuddin, Afroz and Kaiser, Lukasz and others},
  journal = {International Conference on Learning Representations},
  year    = {2021}
}

@article{dao2022flashattention,
  title   = {FlashAttention: Fast and Memory-Efficient Exact Attention with IO-Awareness},
  author  = {Dao, Tri and Fu, Daniel Y. and Ermon, Stefano and Rudra, Atri and Re, Christopher},
  journal = {Advances in Neural Information Processing Systems},
  volume  = {35},
  pages   = {16344--16359},
  year    = {2022}
}

@book{hirose2012complex,
  title     = {Complex-Valued Neural Networks: Advances and Applications},
  author    = {Hirose, Akira},
  year      = {2012},
  publisher = {John Wiley \& Sons}
}

@book{nitta2009complex,
  title     = {Complex-Valued Neural Networks: Utilizing High-Dimensional Parameters},
  author    = {Nitta, Tohru},
  year      = {2009},
  publisher = {Information Science Reference}
}

@article{lim2021temporal,
  title   = {Temporal Fusion Transformers for Interpretable Multi-horizon Time Series Forecasting},
  author  = {Lim, Bryan and Arik, Sercan O. and Loeff, Nicolas and Pfister, Tomas},
  journal = {International Journal of Forecasting},
  volume  = {37},
  number  = {4},
  pages   = {1748--1764},
  year    = {2021}
}

@article{zhou2021informer,
  title   = {Informer: Beyond Efficient Transformer for Long Sequence Time-Series Forecasting},
  author  = {Zhou, Haoyi and Zhang, Shanghang and Peng, Jieqi and Zhang, Shuai and Li, Jianxin and Xiong, Hui and Zhang, Wancai},
  journal = {Proceedings of AAAI},
  volume  = {35},
  number  = {12},
  pages   = {11106--11115},
  year    = {2021}
}

@article{wu2021autoformer,
  title   = {Autoformer: Decomposition Transformers with Auto-Correlation for Long-Term Series Forecasting},
  author  = {Wu, Haixu and Xu, Jiehui and Wang, Jianmin and Long, Mingsheng},
  journal = {Advances in Neural Information Processing Systems},
  volume  = {34},
  pages   = {22419--22430},
  year    = {2021}
}

@article{zhou2022fedformer,
  title   = {FEDformer: Frequency Enhanced Decomposed Transformer for Long-term Series Forecasting},
  author  = {Zhou, Tian and Ma, Ziqing and Wen, Qingsong and Wang, Xue and Sun, Liang and Jin, Rong},
  journal = {International Conference on Machine Learning},
  pages   = {27268--27286},
  year    = {2022}
}

@article{nie2023time,
  title   = {A Time Series is Worth 64 Words: Long-term Forecasting with Transformers},
  author  = {Nie, Yuqi and Nguyen, Nam and Sinthong, Phanwadee and Kalagnanam, Jayant},
  journal = {International Conference on Learning Representations},
  year    = {2023}
}

@article{wu2023timesnet,
  title   = {TimesNet: Temporal 2D-Variation Modeling for General Time Series Analysis},
  author  = {Wu, Haixu and Hu, Tao and Liu, Yong and Zhou, Hang and Wang, Jianmin and Long, Mingsheng},
  journal = {International Conference on Learning Representations},
  year    = {2023}
}

@article{PhasorFlow,
  title   = {PhasorFlow: A Python Library for Unit Circle Based Computing},
  author  = {Sharma, Vasu},
  journal = {arXiv preprint arXiv:2603.15886},
  year    = {2026}
}

@article{sigdel2026phasorflow,
  title   = {PhasorFlow: A Python Library for Unit Circle Based Computing},
  author  = {Sigdel, Dibakar and Panday, Namuna},
  journal = {arXiv preprint arXiv:2603.15886},
  year    = {2026},
  url     = {https://arxiv.org/abs/2603.15886}
}

\end{document}